\documentclass[runningheads]{llncs}
\usepackage{graphicx}
\usepackage{amsmath,amssymb} 
\usepackage{color}
\usepackage{booktabs}
\usepackage{url}

\usepackage
[compatibility=false]
{caption}
\usepackage[list=true]{subcaption}

\begin{document}

\title{Localizing Small Apples in Complex Apple Orchard Environments} 
\titlerunning{Localizing Small Apples in Complex Apple Orchard Environments} 


\author{Christian Wilms \and Robert Johanson \and Simone Frintrop}
%

\authorrunning{C. Wilms et al.} 


\institute{University of Hamburg\\
\email{\{christian.wilms,robert.johanson,simone.frintrop\}@uni-hamburg.de}}

\maketitle

\begin{abstract}
The localization of fruits is an essential first step in automated agricultural pipelines for yield estimation or fruit picking. One example of this is the localization of apples in images of entire apple trees. 
 Since the apples are very small objects in such scenarios, we tackle this problem by adapting the object proposal generation system AttentionMask that focuses on small objects. We adapt AttentionMask by either adding a new module for very small apples or integrating it into a tiling framework. Both approaches clearly outperform standard object proposal generation systems on the MinneApple dataset covering complex apple orchard environments. Our evaluation further analyses the improvement w.r.t. the apple sizes and shows the different characteristics of our two approaches.
\end{abstract}

\section{Introduction}
\label{sec:intro}
Pre-harvest yield estimation is an integral part of agriculture for efficiently planning harvest, transportation, or storing fruits~\cite{anderson2019estimation,hani2020comparative,koirala2019deep,wang2013automated}. Yield estimation usually relies on tedious manual counting in sample areas~\cite{anderson2019estimation,koirala2019deep,wang2013automated}. Such estimations are inaccurate despite additional weather information or historical data~\cite{koirala2019deep}. To support manual yield estimation, computer vision approaches were proposed to automatically localize fruits~\cite{koirala2019deepReview}. This improves yield estimation~\cite{wang2013automated} or automated fruit picking~\cite{yu2019fruit}, among other applications.

Fruit localization is an object detection, instance segmentation, or object proposal generation task. The exact formulation depends on the annotation type and the need for classification. Since the advent of deep learning, CNN-based systems have led to strong improvements on all three tasks~\cite{he2017mask,redmon2016you,ren2016faster,WilmsFrintropACCV2018}. Therefore, some of these systems were also applied to fruit localization~\cite{bargoti2017deep,koirala2019deep,mai2018faster,sa2016deepfruits,stein2016image}.

\begin{figure}[t]
\centering
\subcaptionbox{COCO dataset\label{fig:minneAppleVsCOCO_coco}}{\includegraphics[height=4cm]{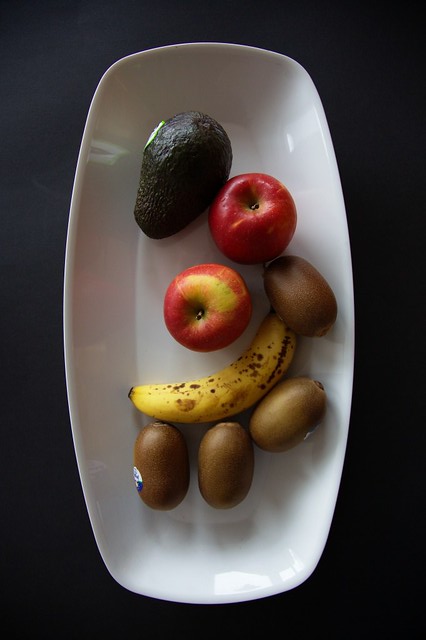}~~~~~\includegraphics[height=4cm]{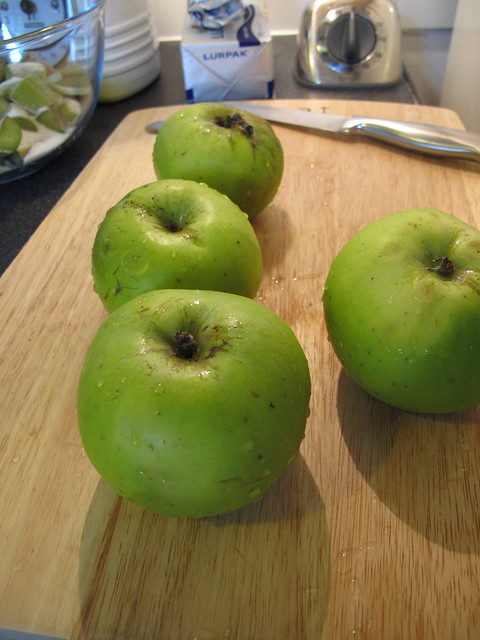}}%
\hfill
\subcaptionbox{MinneApple dataset\label{fig:minneAppleVsCOCO_minneApple}}{\includegraphics[height=4cm]{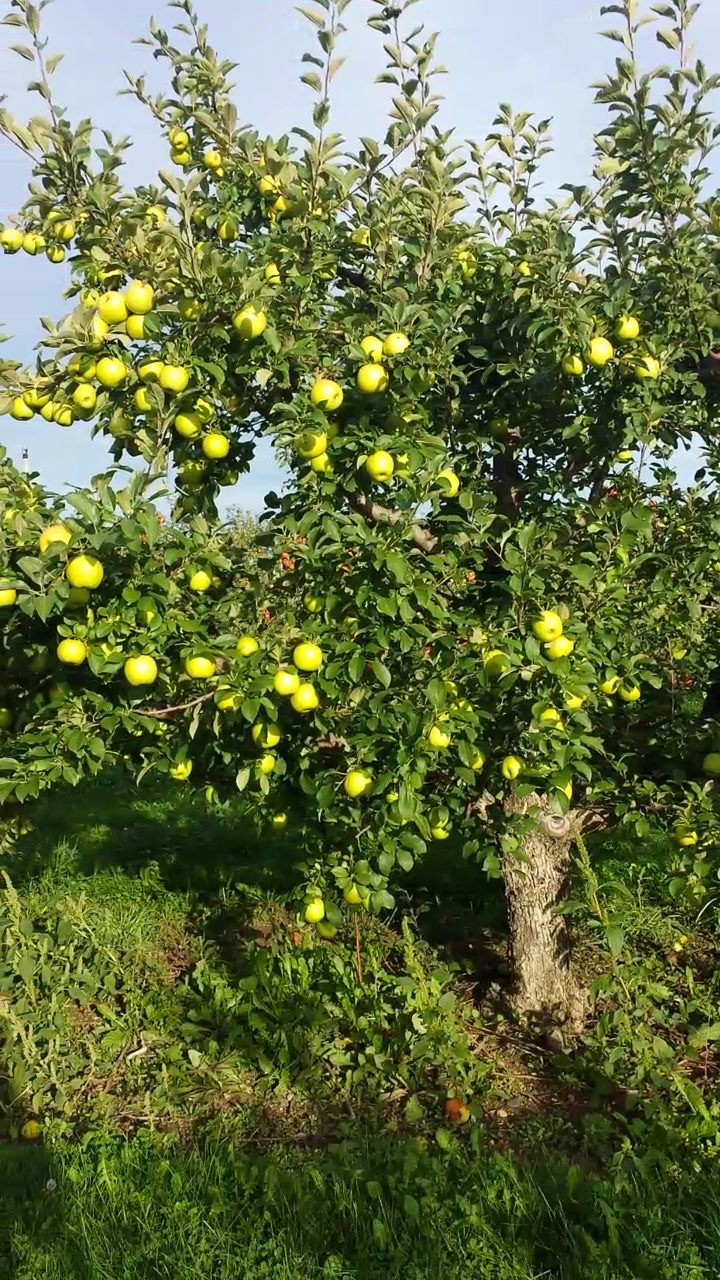}~~~~~\includegraphics[height=4cm]{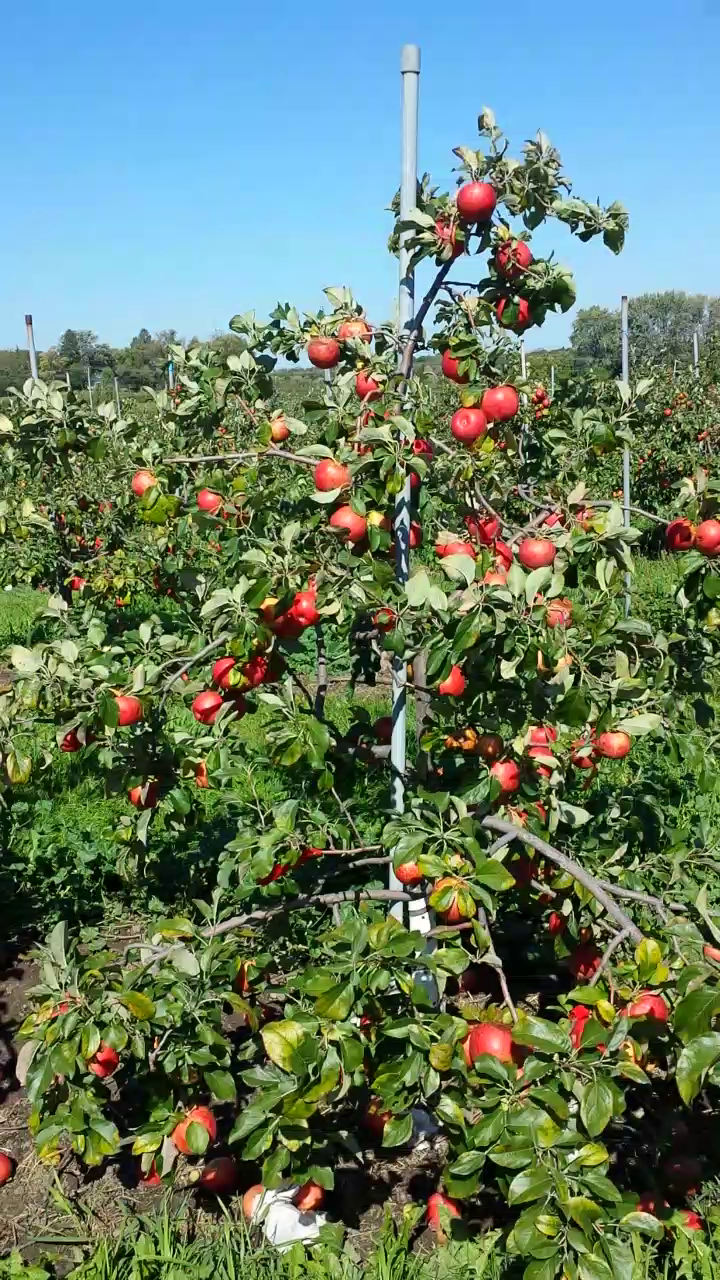}}%
\caption{Example image containing apples from the COCO dataset~\cite{lin2014microsoft}~(\subref{fig:minneAppleVsCOCO_coco}) and the MinneApple dataset~\cite{hani2020minneapple}~(\subref{fig:minneAppleVsCOCO_minneApple}). While the COCO images depict simple scenes, the MinneApple images include challenges like occlusions, clutter, and small apples.
}
\label{fig:minneAppleVsCOCO}
\end{figure}

We focus on the subarea of apple localization in orchard environments. Apples are a common object class in computer vision datasets as well~\cite{lin2014microsoft}. However, unlike apples in the agricultural context, in datasets like COCO~\cite{lin2014microsoft}, apples are primarily presented in simple environments and unobstructed views (see Fig.~\ref{fig:minneAppleVsCOCO_coco}). In contrast, the MinneApple dataset~\cite{hani2020minneapple} for agricultural applications, covers apples in their complex natural orchard environment. Apples in apple orchards are harder to localize due to occlusions and clutter. Moreover, the apples are very small when covering an entire apple tree in one image, as the examples in Fig.~\ref{fig:minneAppleVsCOCO_minneApple} demonstrate. This makes the localization of apples or objects in general difficult~\cite{koirala2019deep,liu2020deep,WilmsFrintropACCV2018}. 

In this paper, we adapt the object proposal generation system AttentionMask~\cite{WilmsFrintropACCV2018} to apple localization. Applying an object proposal generation system to this task is reasonable since no classification is required. With its strong performance on localizing small objects, AttentionMask is well-suited for apple localization in orchard environments. From AttentionMask, we derive two variations to improve the localization of very small apples based on a new module for such apples and a tiling approach. Our evaluation reveals an improved performance for both variations compared to object proposal generation methods. Furthermore, a characterization of differences between the variations allows a task-specific choice.


\section{Related Work}
\label{sec:relWork}
This section briefly reviews important work in the related fields of fruit localization and object proposal generation.

\subsection{Fruit Localization}
\label{sec:relWork_fruit}
For the localization of fruits, several approaches were proposed. An overview of recent CNN-based methods is given by~\cite{koirala2019deepReview}, while~\cite{gongal2015sensors} present older hand-crafted methods. Generally, CNN-based methods mostly adapt standard object detection systems~\cite{bargoti2017deep,koirala2019deep,mai2018faster,sa2016deepfruits,stein2016image} like Faster R-CNN~\cite{ren2016faster} or standard instance segmentation systems~\cite{yu2019fruit} like Mask R-CNN~\cite{he2017mask}. For instance, \cite{sa2016deepfruits}~adapt Faster R-CNN to handle RGB and near-infrared images for localizing sweet peppers, while \cite{mai2018faster} augment Faster R-CNN with additional branches for generating smaller object proposals in almond localization. Similarly, \cite{koirala2019deep}~adapt YOLO~\cite{redmon2016you} by reducing the model size and aggregating features for mango localization. Finally, \cite{yu2019fruit}~utilize Mask R-CNN to locate strawberries and add a module to estimate the picking point.

In contrast to those methods, we adapt an object proposal generation system without unnecessary classification modules. Unlike object detection-based systems, we locate fruits based on pixel-precise masks.

\subsection{Object Proposal Generation}
\label{sec:relWork_opg}
The aim of object proposal generation is to propose a ranked set of object candidates as boxes or pixel-precise masks. Since the seminal work of~\cite{alexe2010object}, systems based on hand-crafted features~\cite{Hosang2015PAMI} and CNNs~\cite{Hu2017-fastmask,Pinheiro2015-deepmask,Pinheiro2016-sharpmask,WilmsFrintropACCV2018,WilmsFrintropICPR2020,WilmsFrintropIVC2021} have been proposed. \cite{Pinheiro2015-deepmask} introduce one of the first CNN-based approaches. It extracts crops of an image pyramid and generates an objectness score for ranking and a segmentation mask per crop. This is achieved by two heads on top of a backbone network. Since applying the backbone on overlapping crops of the input image is time-consuming, \cite{Hu2017-fastmask}~propose FastMask and move the pyramid inside the network. Hence, a feature pyramid inside the CNN is created. Subsequently, windows are extracted from the pyramid for segmentation and objectness scoring. Improving the efficiency and the results on small objects, AttentionMask~\cite{WilmsFrintropACCV2018} utilizes the concept of visual attention and extracts only relevant windows from the feature pyramid. Further improving the adherence of AttentionMask proposals to the object boundaries, \cite{WilmsFrintropICPR2020,WilmsFrintropIVC2021}~introduce a superpixel-based refinement. 

Different from the discussed approaches, we propose a system that focuses on small and very small apples as objects. We discuss AttentionMask in more detail in Sec.~\ref{sec:attMask}, since it serves as the baseline system for our proposed apple localization systems.

\section{Baseline System: AttentionMask}
\label{sec:attMask}
\begin{figure}[t]
    \centering
    \includegraphics[width=\textwidth]{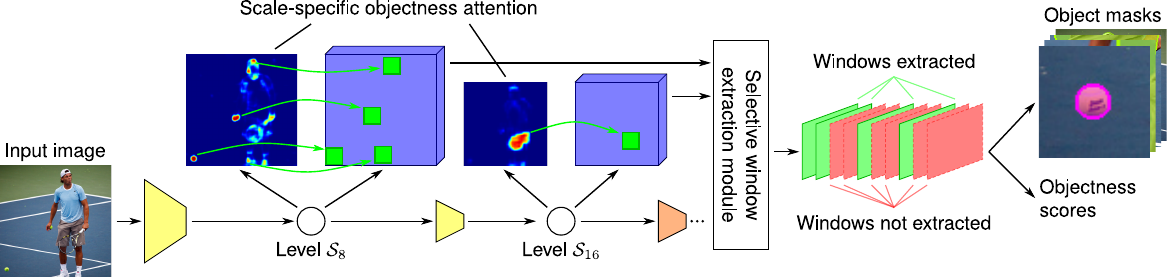}
    \caption{Abstract system figure of AttentionMask. The first part of the split backbone (yellow) processes the input image, yielding the base pyramid level $\mathcal{S}_8$. Further pyramid levels are created using the second part of the backbone or residual necks (orange). Per pyramid level, an attention map focuses the extraction of windows on relevant areas (red areas in heat maps). A segmentation and an objectness score are generated for each extracted window (green windows). Note that the input image is cropped for clarity. The figure is based on~\cite{WilmsFrintropACCV2018}.}\label{fig:abstractSysFig}
\end{figure}

As outlined above, we use the object proposal generation system AttentionMask~\cite{WilmsFrintropACCV2018} as the baseline system for our apple localization approaches. AttentionMask follows FastMask~\cite{Hu2017-fastmask} and generates a feature pyramid inside the CNN based on a ResNet~\cite{he2016deep} backbone. The pyramid is partially visualized in the center of Fig.~\ref{fig:abstractSysFig}. Note that only the first two pyramid levels are depicted for simplicity. The base level of the pyramid is the output of the backbone's first part. It is a feature map downscaled by a factor of 8 w.r.t. the input image. Hence, we coin this pyramid level $\mathcal{S}_8$. Further pyramid levels up to $\mathcal{S}_{128}$ are created using the remainder of the backbone and residual neck modules~\cite{Hu2017-fastmask}. From this pyramid, $10 \times 10$ windows are extracted. The extraction of fixed size windows across the different levels allows AttentionMask to locate objects of different sizes. Small objects are localized at $\mathcal{S}_8$, while large objects are localized at $\mathcal{S}_{128}$. However, unlike FastMask, a scale-specific objectness attention map (see heat maps in Fig.~\ref{fig:abstractSysFig}) is created per pyramid level using Scale-specific Objectness Attention Modules (SOAMs). The SOAMs utilize the concept of visual attention and highlight parts of each pyramid levels' feature map that contain objects of the relevant size. Thus, only a few windows (indicated by green windows in Fig.~\ref{fig:abstractSysFig}) are extracted from the feature pyramid. This selective extraction allows the addition of the base feature map $\mathcal{S}_8$ and makes AttentionMask more efficient.

For each extracted window, a segmentation is created to separate the object from the background (top right in Fig.~\ref{fig:abstractSysFig}). Additionally, an objectness score is generated per window as a confidence score (bottom right in Fig.~\ref{fig:abstractSysFig}). Note that the entire system is learned end-to-end.

\section{Method}
\label{sec:method}
Our two proposed apple localization systems that focus on locating small and very small apples in complex orchard environments are based on the previously introduced AttentionMask~\cite{WilmsFrintropACCV2018}. As discussed earlier, AttentionMask achieves state-of-the-art results on object proposal generation and already focuses on small objects. Additionally, AttentionMask shows a strong performance in complex real-world environments~\cite{WilmsEtAlICPR2020}. Hence, AttentionMask is a reasonable baseline system to localize small and  very small apples in complex orchard environments. The subsequent sections present our extensions to AttentionMask. First, in Sec.~\ref{sec:method_attMask-4-16} we propose the AttentionMask variation AttentionMask$_4^{16}$ with a pyramid level for localizing very small apples. In contrast, Sec.~\ref{sec:method_tiles} integrates AttentionMask in a tiling framework to increase the relative size of the apples in the input image.

\subsection{Extended Feature Pyramid}
\label{sec:method_attMask-4-16}
\begin{figure}[t]
    \centering
    \includegraphics[width=\textwidth]{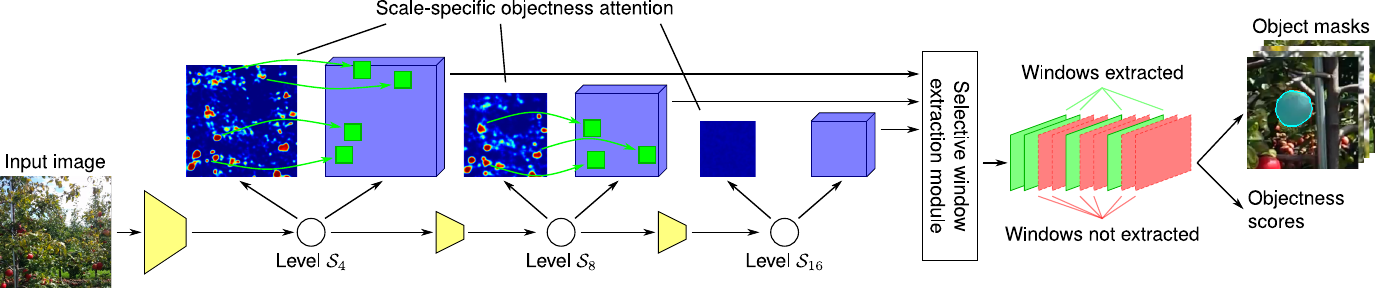}
    \caption{Abstract system figure of our proposed AttentionMask$_4^{16}$. Compared to to Fig.~\ref{fig:abstractSysFig}, the first part of the backbone is split again, leading to three backbone parts (yellow) and a new pyramid level $\mathcal{S}_4$. No residual necks and no further pyramid levels after $\mathcal{S}_{16}$ exist. Note that the input image is cropped for clarity. The figure is based on~\cite{WilmsFrintropACCV2018}.}\label{fig:attMask_4_16}
\end{figure}

Our first apple localization method adapts the original AttentionMask in two ways. First, we split the first part of the backbone network as visible in Fig.~\ref{fig:attMask_4_16} to extract the final features of the ResNet's \textit{conv2} stage. These features are only downsampled by a factor of 4 w.r.t. the input images. Thus, more details are preserved compared to the features from the \textit{conv3} stage used for $\mathcal{S}_8$. The new feature map is the new base level ($\mathcal{S}_4$) of the feature pyramid for very small apples (see Fig.~\ref{fig:attMask_4_16}). Compared to standard AttentionMask, the minimal side length of a localizable apple is reduced by a factor of $\frac{1}{2}$. A drawback of using \textit{conv2} features is the reduced semantic richness compared to \textit{conv3} or \textit{conv4} features. However, since very small apples are simple objects without a visible texture (see Fig.~\ref{fig:minneAppleVsCOCO_minneApple}), it is reasonable to assume that \textit{conv2} features are sufficient for localizing very small apples.

The second modification of AttentionMask is removing all pyramid levels above $\mathcal{S}_{16}$. In its standard formulation, AttentionMask consists of eight pyramid levels: $\mathcal{S}_{8}$ - $\mathcal{S}_{128}$. Since the later pyramid levels locate large objects, the later layers are unnecessary for the localization of the small and very small apples. Hence, we can remove these pyramid levels as they increase the runtime and induce false positives. The rest of AttentionMask, including the training regime, is unchanged and generates pixel-precise apple proposals as visualized in Fig.~\ref{fig:attMask_4_16}. Since we only use the pyramid levels $\mathcal{S}_{4}$ - $\mathcal{S}_{16}$, we follow the naming conventions in~\cite{WilmsFrintropACCV2018} and coin this approach AttentionMask$_4^{16}$.

\subsection{Tiled Processing}
\label{sec:method_tiles}
Our second approach for localizing small and very small apples, coined Tiled AttentionMask, embeds AttentionMask into a tiling framework. We extract 26 overlapping tiles from the input image, reducing the input image size from $1280 \times 720$ (MinneApple dataset) to $320\times 240$. Since an input image is rescaled to a predefined size in AttentionMask, the relative size of the apples is essential. Thus, effectively each tile will be upsampled before being processed by the system leading to a larger representation of small apples. Consequentially, after subsampling in the backbone, small apples are still visible in the feature pyramid's base level. We use this tiling approach during training and testing similar to~\cite{bargoti2017deep,koirala2019deep}.

The architecture of AttentionMask is unchanged, while the learning rate is lowered to 0.00007, adapting to the new data. For combining the results of the 26 tiles in testing, we merge all proposals resulting from one original image. On this set of proposals we apply non-maximum suppression (NMS) and rank the proposals based on the objectness scores. Finally, we keep the first 100 proposals.

\section{Evaluation}
\label{sec:eval}
For evaluating the effect of our proposed approaches for apple localization, we use a standard object proposal evaluation pipeline~\cite{Hu2017-fastmask,Pinheiro2015-deepmask,Pinheiro2016-sharpmask,WilmsFrintropACCV2018,WilmsFrintropICPR2020,WilmsFrintropIVC2021} with the MinneApple dataset~\cite{hani2020minneapple}. Thus, we use Average Recall (AR)~\cite{Hosang2015PAMI} as the evaluation measure. AR takes a predefined amount of proposals per image and determines how many annotated objects are localized and how well they are localized. We report AR for the first 10 (AR@10) and the first 100 (AR@100) proposals. Additionally, we follow~\cite{lin2014microsoft} and report AR for different absolute sizes of objects. We use two standard size categories M as well as S, and add a third category (XS) to account for the large number of very small objects~(51\% in the MinneApple dataset). An annotated object fits category M if its area is larger than $32^2$ pixels. Category S covers objects between $32^2$ pixels and $22.5^2$ pixels, while XS covers all objects smaller than $22.5^2$ pixels. 

We use the MinneApple dataset~\cite{hani2020minneapple} for training and testing. The MinneApple dataset contains 41325 manually annotated apples on apple trees in complex orchard environments across 1000 images.  We use the 670 images that feature publicly available annotations. Thus, we split these images into 600 training images, 40 validation images, and 30 test images. All systems are trained on the 600 training images, while all reported results were generated on our test images. 

We compare our proposed approaches AttentionMask$_4^{16}$ and Tiled AttentionMask to standard AttentionMask~\cite{WilmsFrintropACCV2018} and standard FastMask~\cite{Hu2017-fastmask}, which achieves competitive results in object proposal generation. Additionally, we apply our tiling framework to FastMask and also evaluate Tiled FastMask. A comparison to the systems in Sec.~\ref{sec:relWork_fruit} is impossible as no code is publicly available.

\subsection{Quantitative Results}
\label{sec:quanRes}
The quantitative results are presented in Tab.~\ref{tab:apples_ar}. Across all apple sizes (AR@100), our proposed Tiled AttentionMask outperforms all other systems, including At\-ten\-tion\-Mask$_4^{16}$. The improvement w.r.t. standard AttentionMask is 70.8\%, while At\-ten\-tion\-Mask$_4^{16}$ is still outperformed by 33.4\%. Yet, AttentionMask$_4^{16}$ shows a substantial improvement compared to standard AttentionMask (+28.0\%). On AR@10, AttentionMask$_4^{16}$ outperforms Tiled AttentionMask. Standard FastMask is unable to locate any apple, since FastMask's feature pyramid lacks a pyramid level $\mathcal{S}_8$. After applying our tiling, FastMask generates competitive results. However, Tiled AttentionMask still outperforms Tiled FastMask by 17.2\% (AR@100) due to the pyramid level $\mathcal{S}_8$ that allows locating smaller apples.

\begin{table}[t]
\centering
\caption{Quantitative results of standard AttentionMask~\cite{WilmsFrintropACCV2018}, standard FastMask~\cite{Hu2017-fastmask} as well as the proposed AttentionMask$_4^{16}$, Tiled FastMask, and Tiled AttentionMask in terms of various Average Recall (AR) measures on our MinneApple test dataset. AR$^{XS}$, AR$^{S}$, and AR$^{M}$ denote Average Recall results on very small, small, and medium apples.}
\label{tab:apples_ar}
\begin{tabular}{lcccccc}
\toprule
System & AR@10$\uparrow$ & AR@100$\uparrow$ & AR$^{XS}$@100$\uparrow$ & AR$^{S}$@100$\uparrow$ & AR$^{M}$@100$\uparrow$ \\ \hline 
AttentionMask~\cite{WilmsFrintropACCV2018} &  0.071 & 0.243 & 0.126 & 0.336 & 0.337 \\  
FastMask~\cite{Hu2017-fastmask} &  0.000 & 0.000 & 0.000 & 0.000& 0.000 \\ \midrule 
AttentionMask$_4^{16}$ &  \textbf{0.099} & 0.311 & 0.153 & 0.386 & \textbf{0.570}  \\ \midrule 
Tiled FastMask &  0.061 & 0.354 & 0.284 & 0.423 & 0.374 \\ 
Tiled AttentionMask &  0.073 & \textbf{0.415} & \textbf{0.294} & \textbf{0.500} & 0.549  \\ \bottomrule 
\end{tabular}
\end{table}

The size-specific results show a clear difference between standard AttentionMask and the two proposed approaches. While the results for small and medium apples (AR$^S$@100 and AR$^M$@100) are better by at least 14.9\%, compared to standard AttentionMask, the improvement for very small apples is substantially bigger. AttentionMask$_4^{16}$ outperforms standard AttentionMask by 21.4\% on very small apples (AR$^{XS}$@100). Tiled AttentionMask even reaches an improvement of 133\% w.r.t. standard AttentionMask. This clearly shows the substantial advantage of the proposed changes to AttentionMask for locating very small apples. Despite the improved results utilizing the tiling framework, it is worth emphasizing that the GPU runtime is increased by a factor of 15 compared to AttentionMask$_4^{16}$. Hence, AttentionMask$_4^{16}$ is preferable for efficient performance, while Tiled AttentionMask generates the best overall localization results.

\subsection{Qualitative Results}
\label{sec:qualRes}
\begin{figure}[t]
\centering
\begin{tabular}{cccccc}
\includegraphics[width = .15\linewidth]{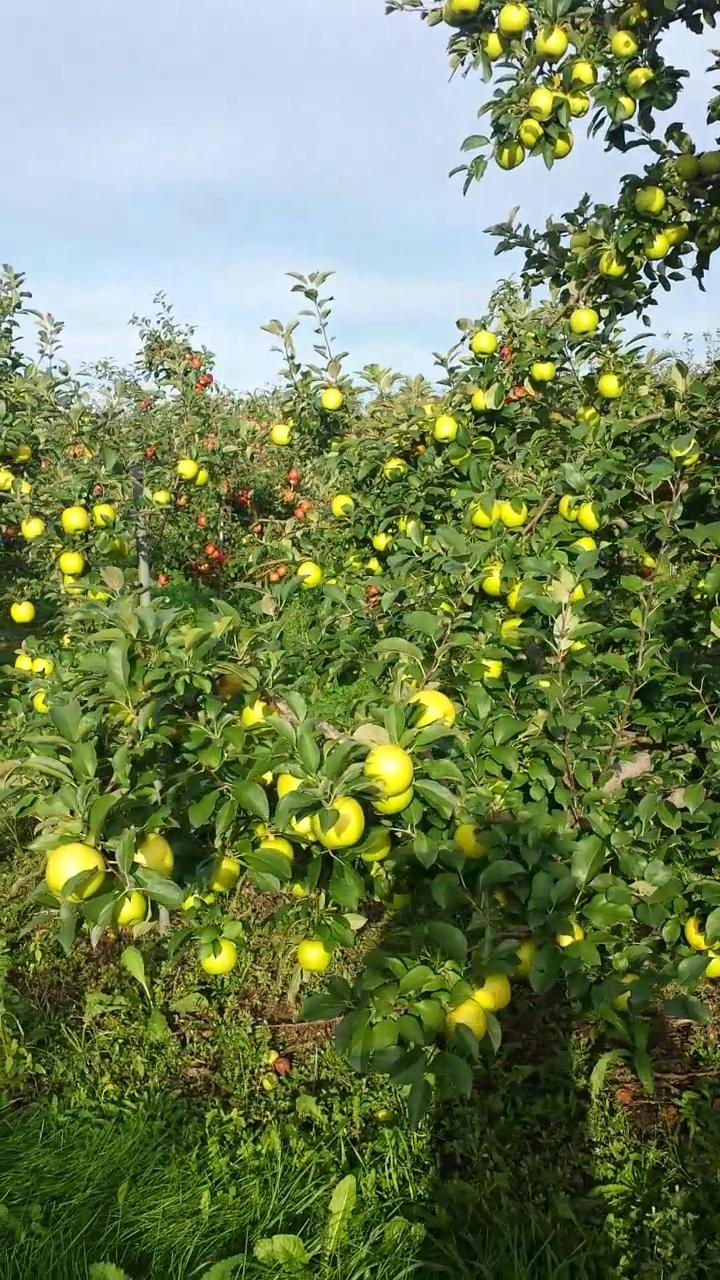} &
\includegraphics[width = .15\linewidth]{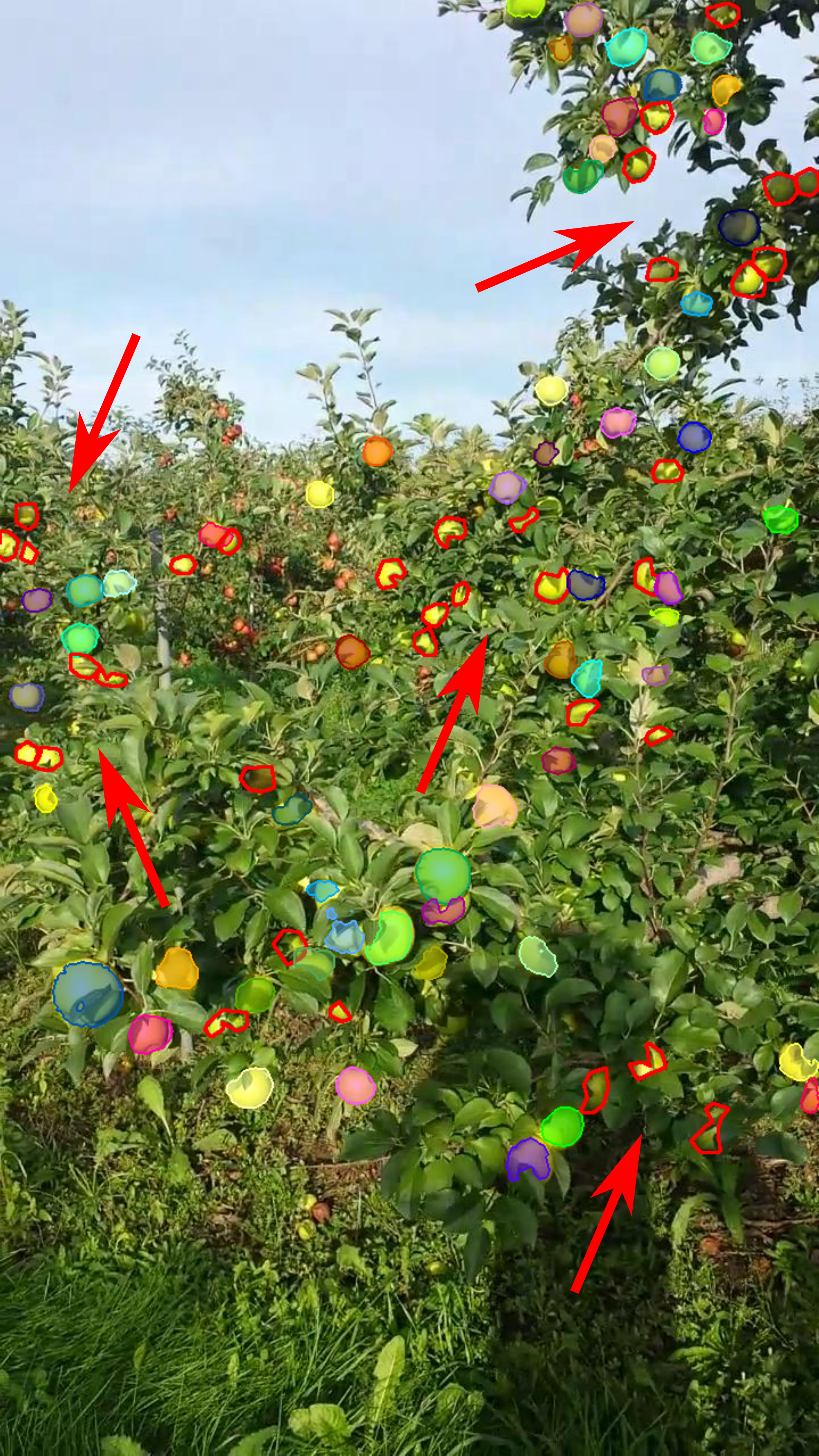} &
\includegraphics[width = .15\linewidth]{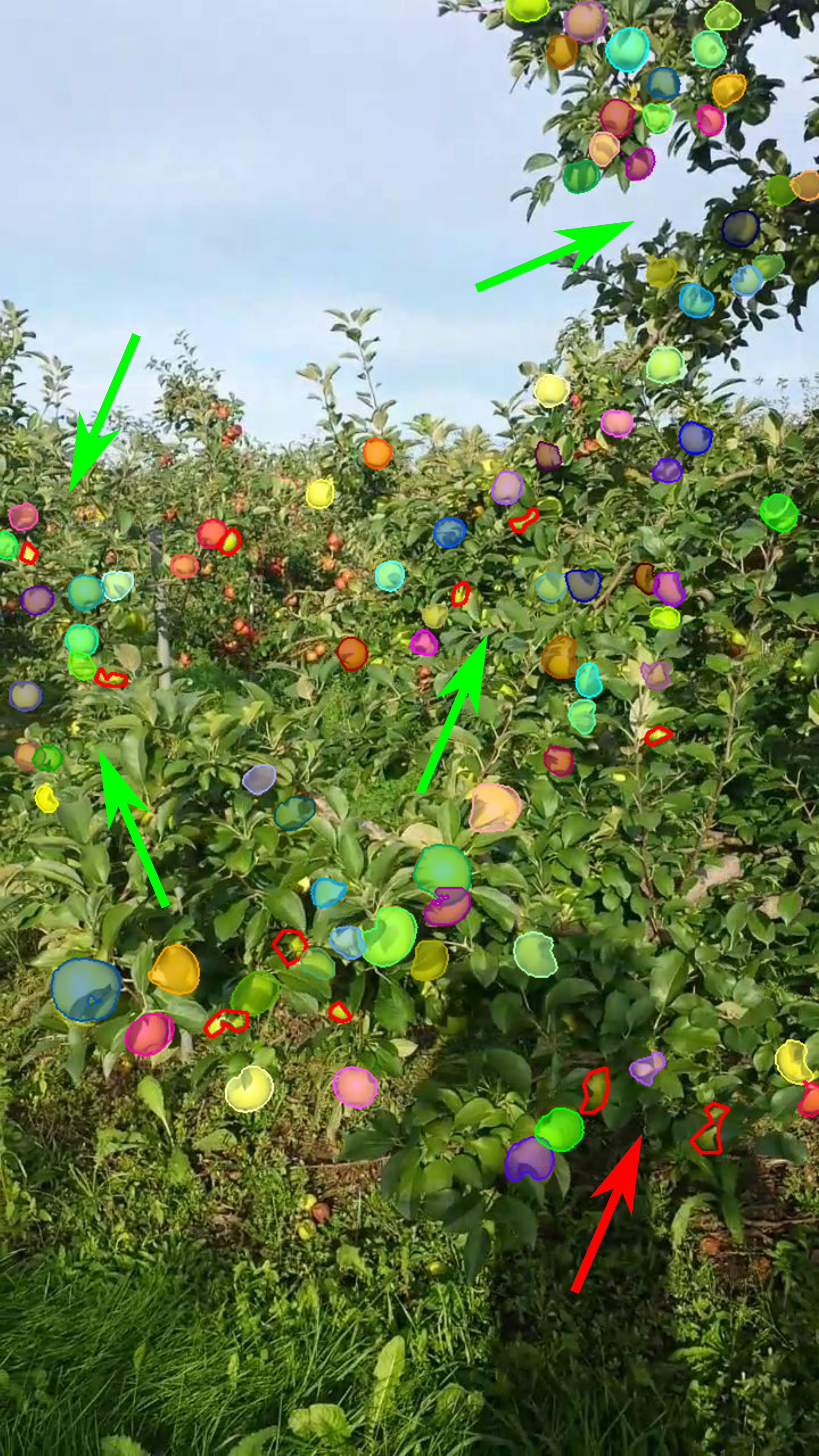} &
\includegraphics[width = .15\linewidth]{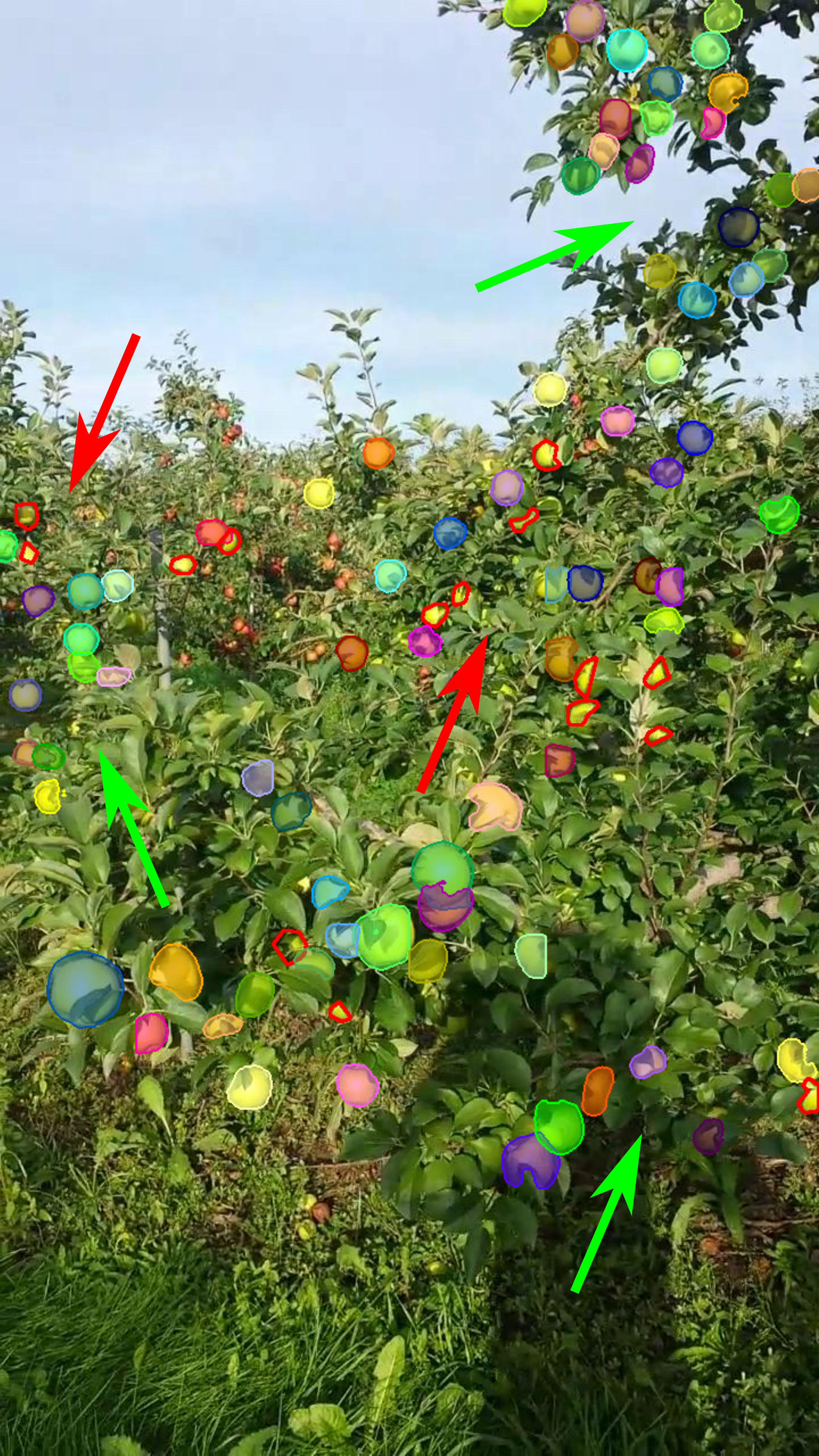} &
\includegraphics[width = .15\linewidth]{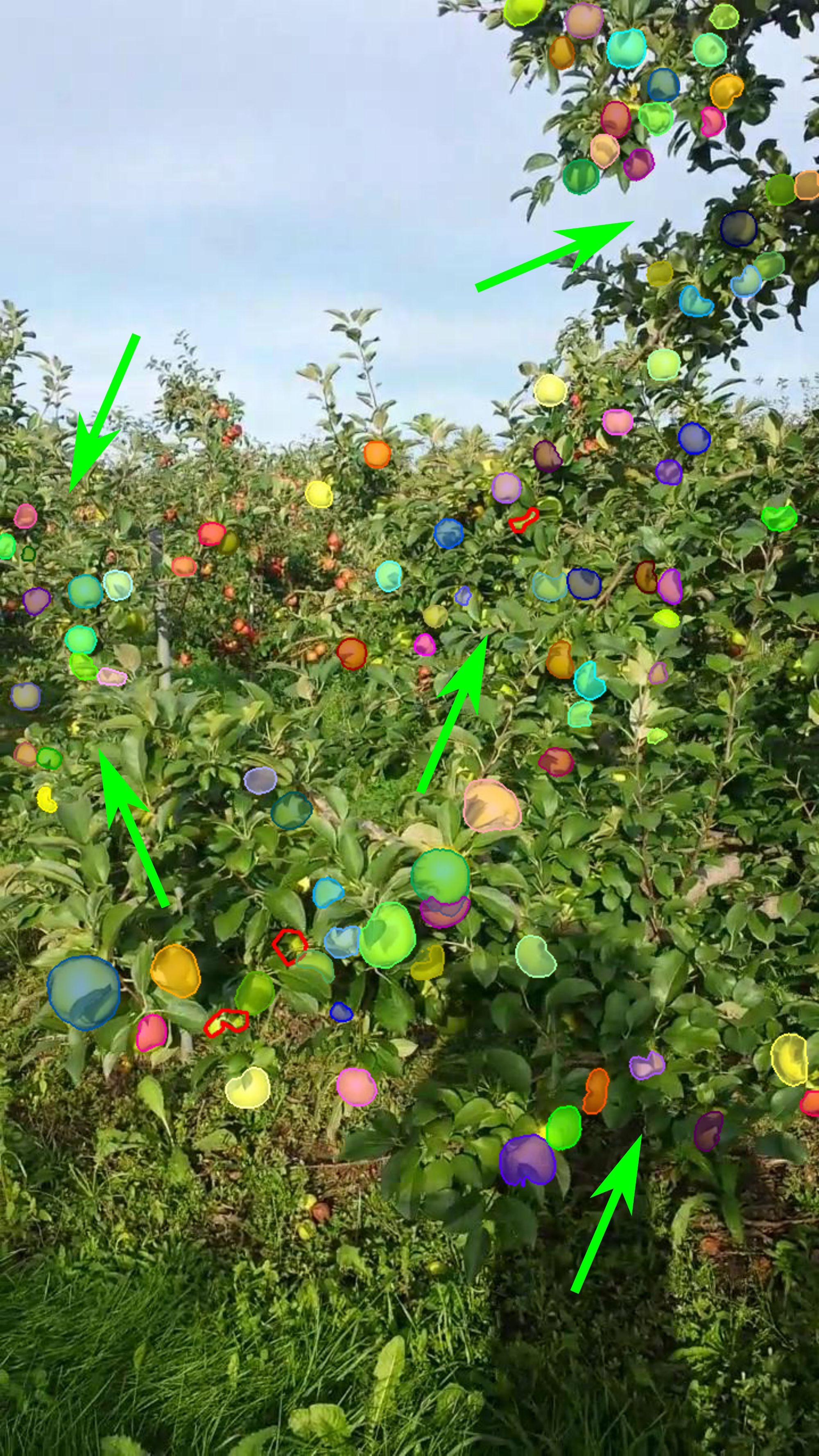} &
\includegraphics[width = .15\linewidth]{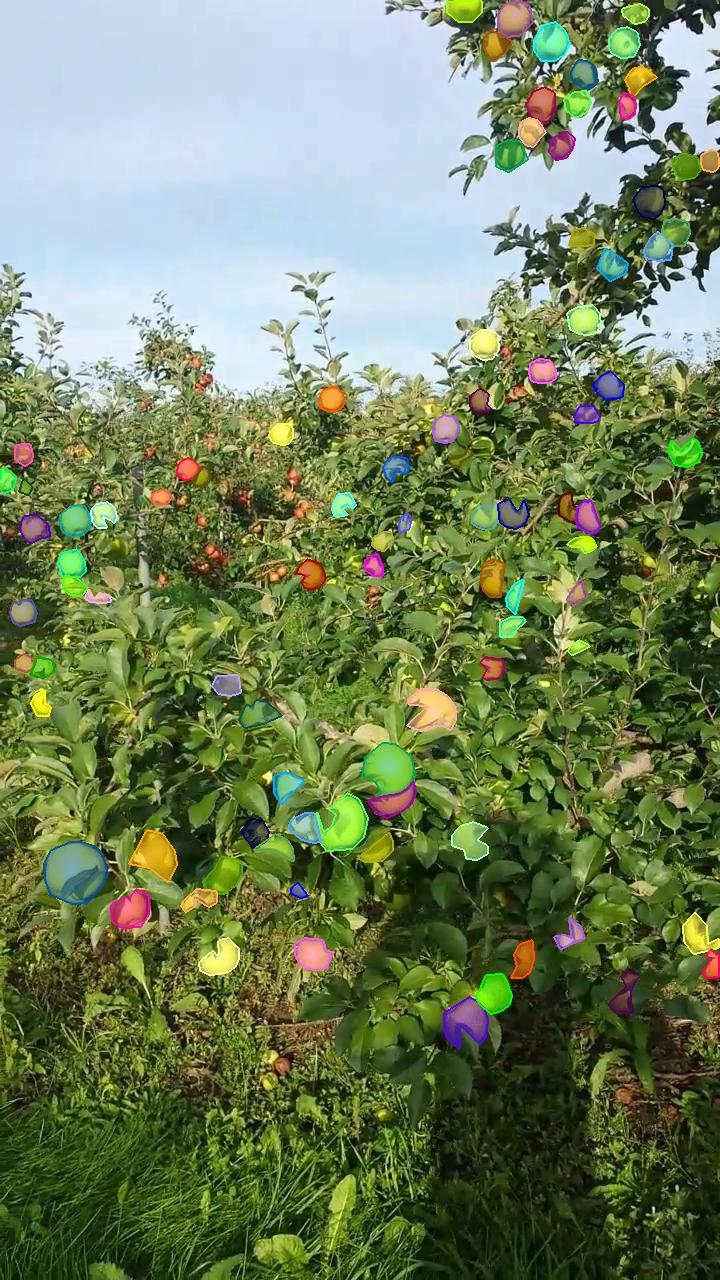}\\
\includegraphics[width = .15\linewidth]{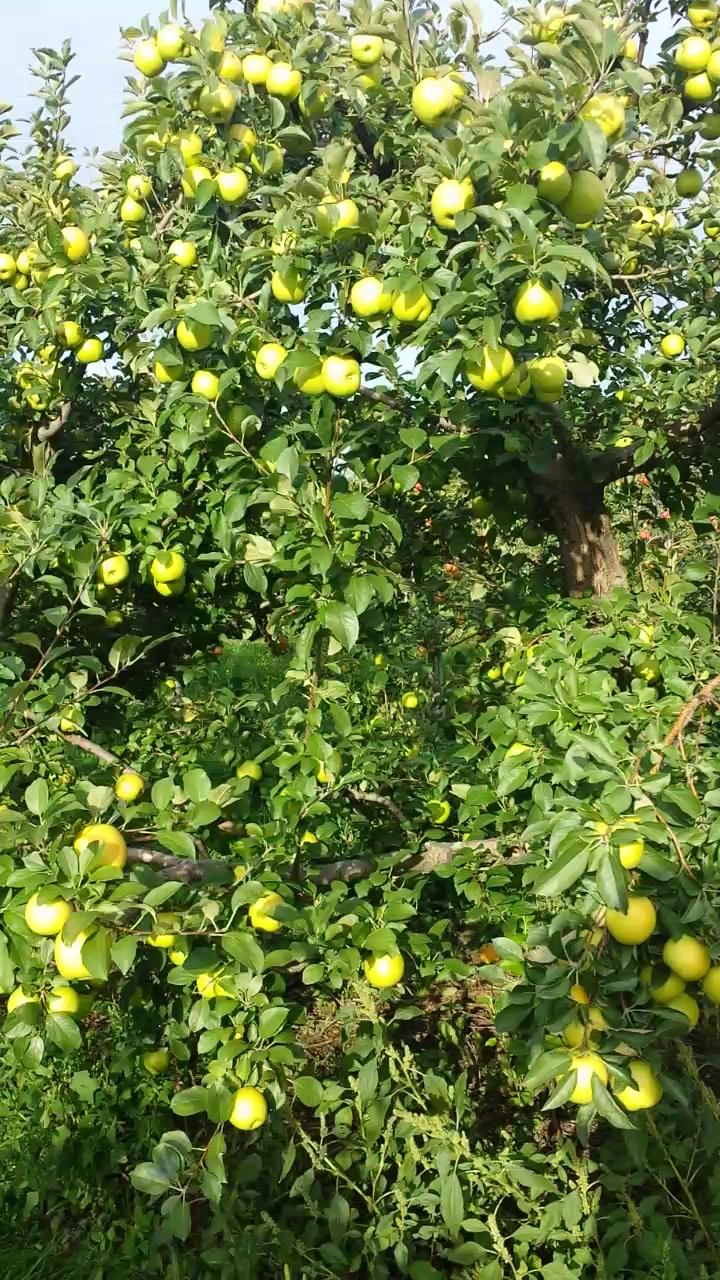} &
\includegraphics[width = .15\linewidth]{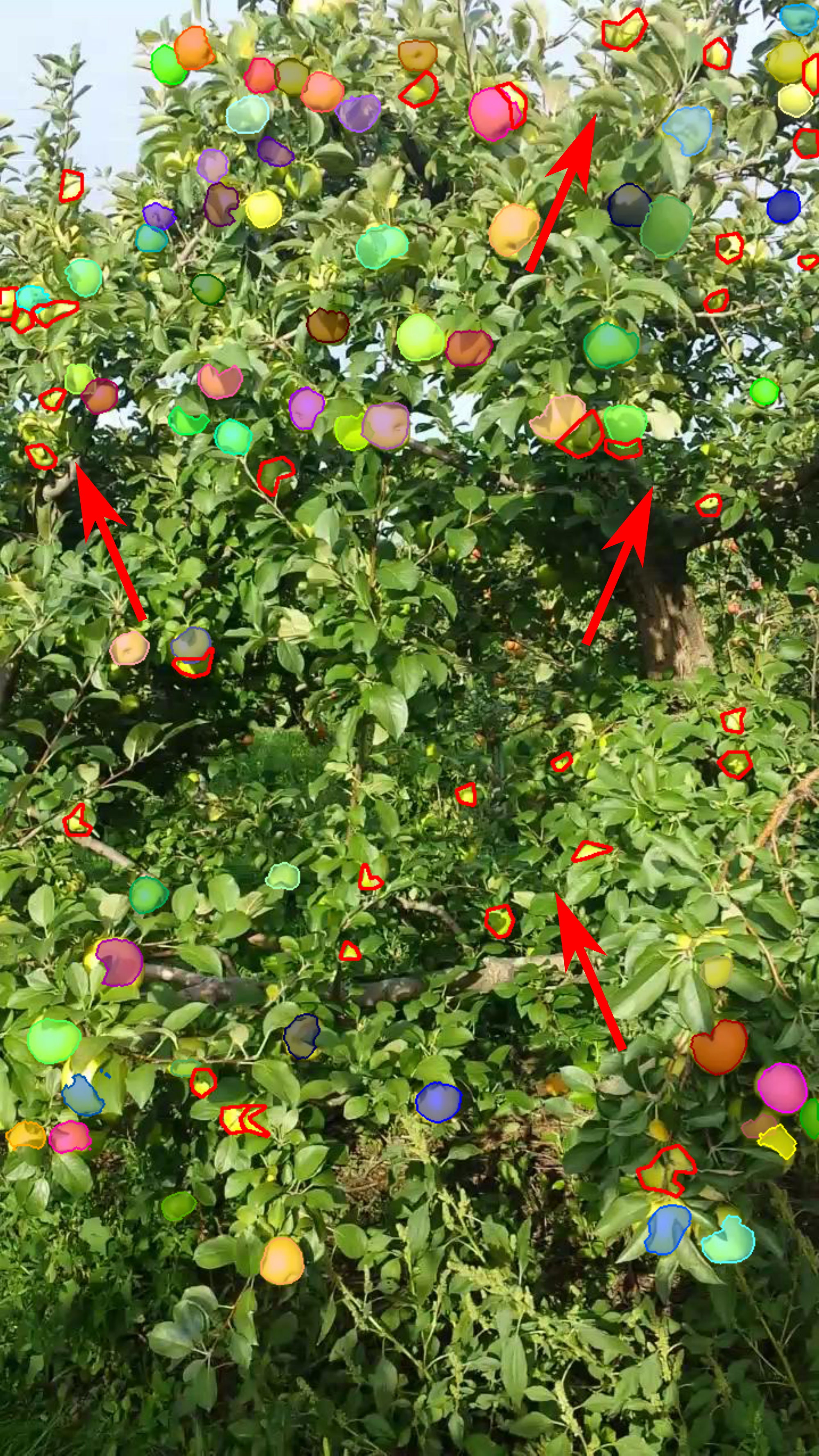} &
\includegraphics[width = .15\linewidth]{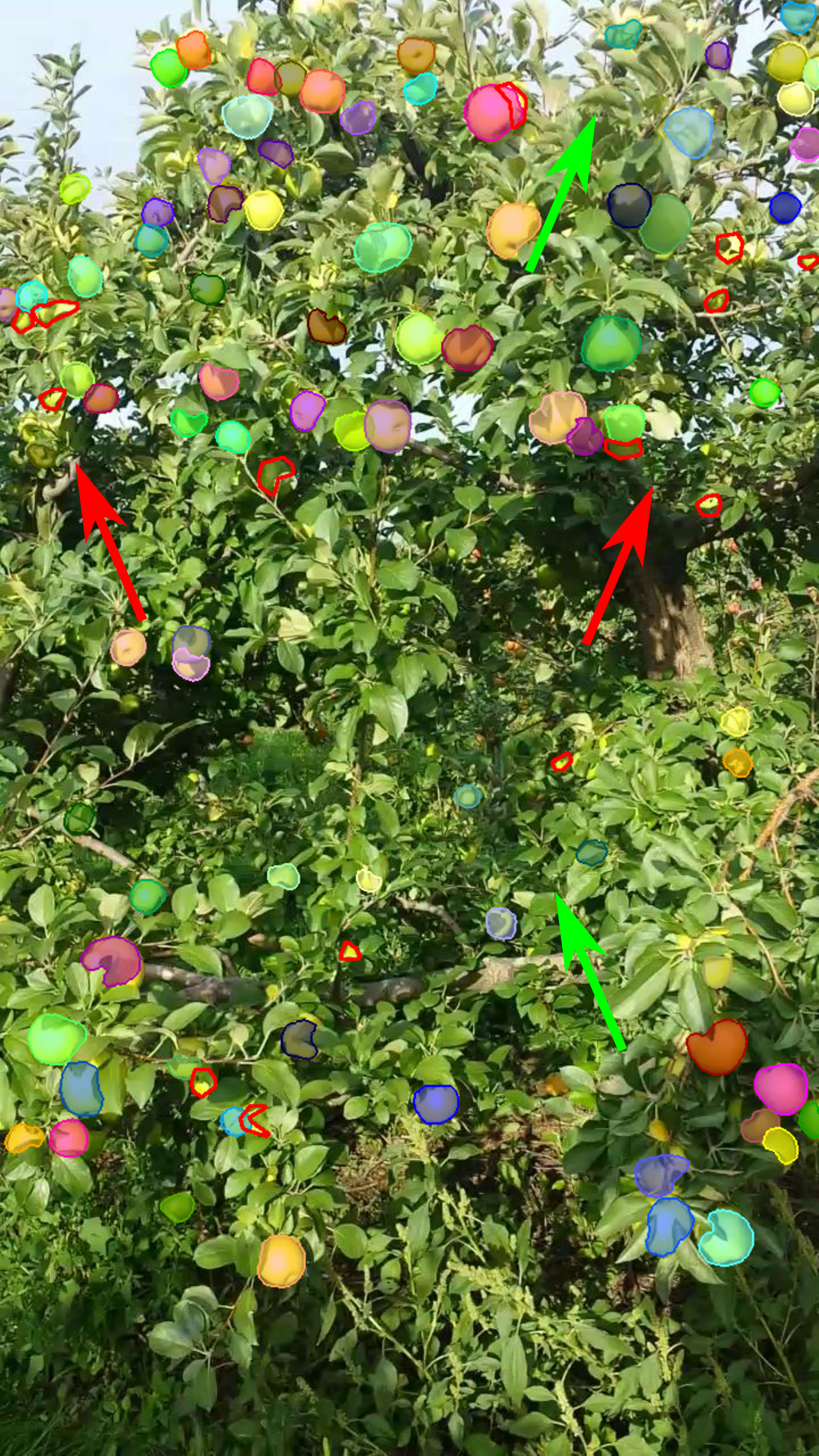} &
\includegraphics[width = .15\linewidth]{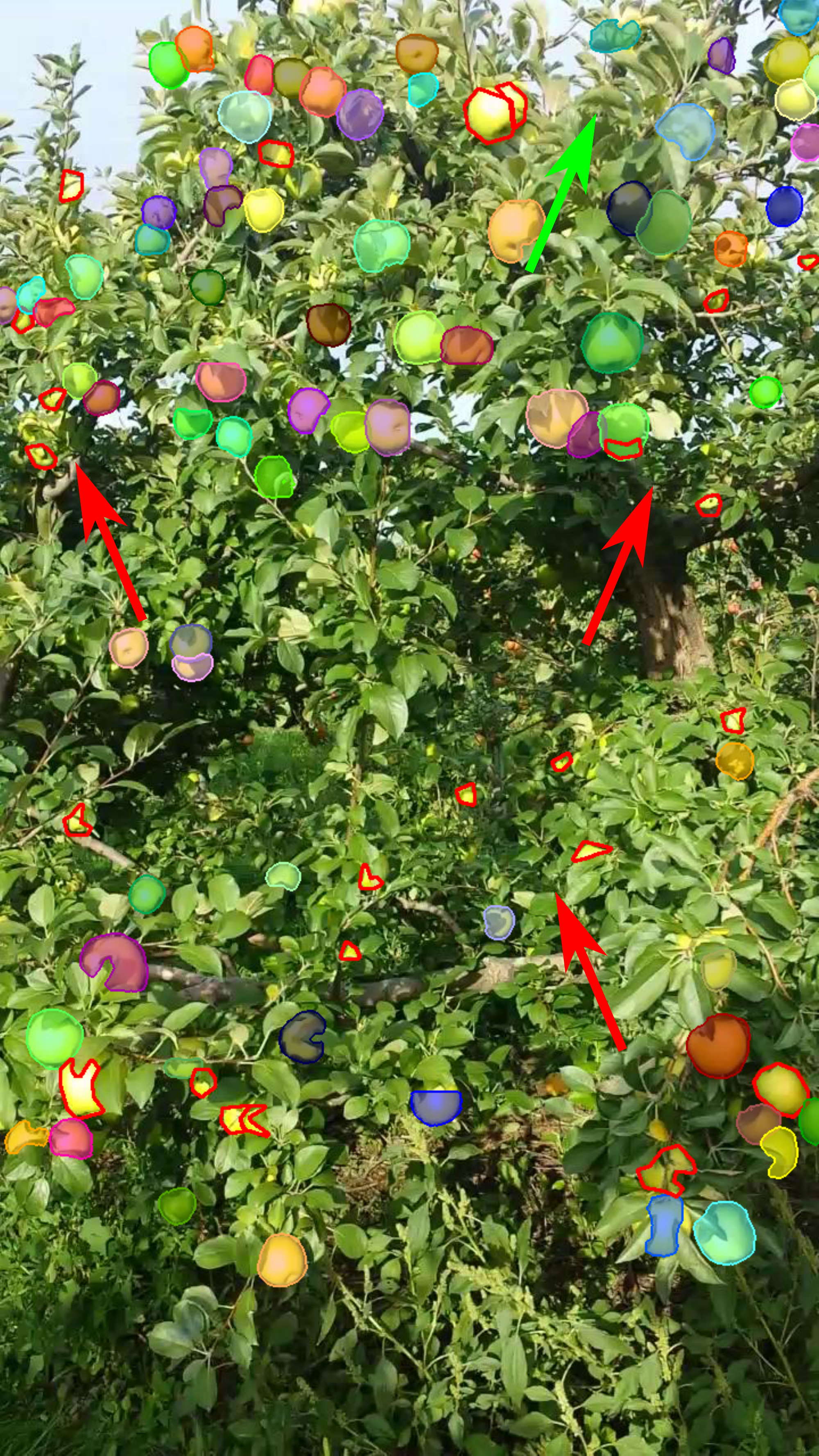} &
\includegraphics[width = .15\linewidth]{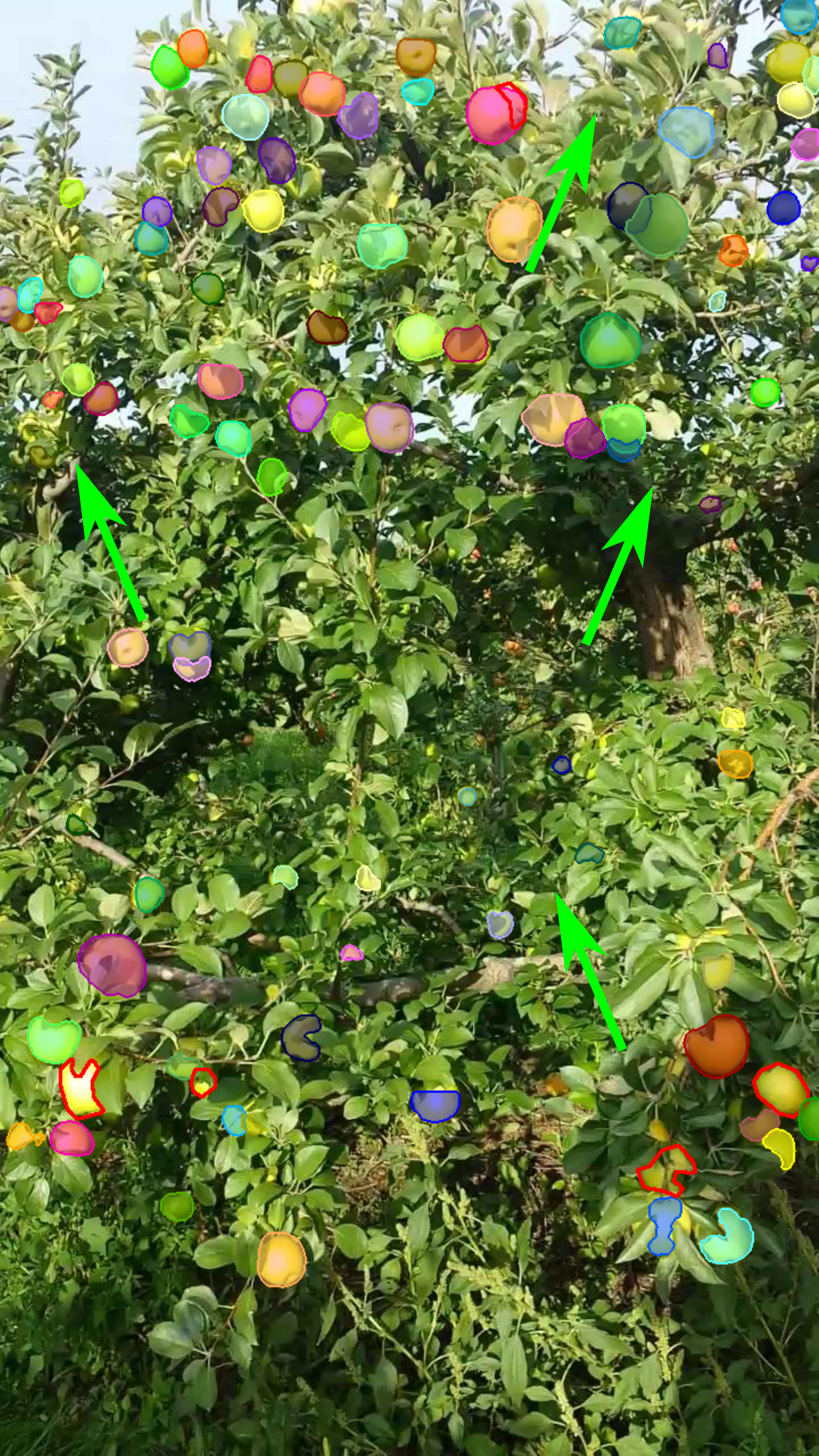} &
\includegraphics[width = .15\linewidth]{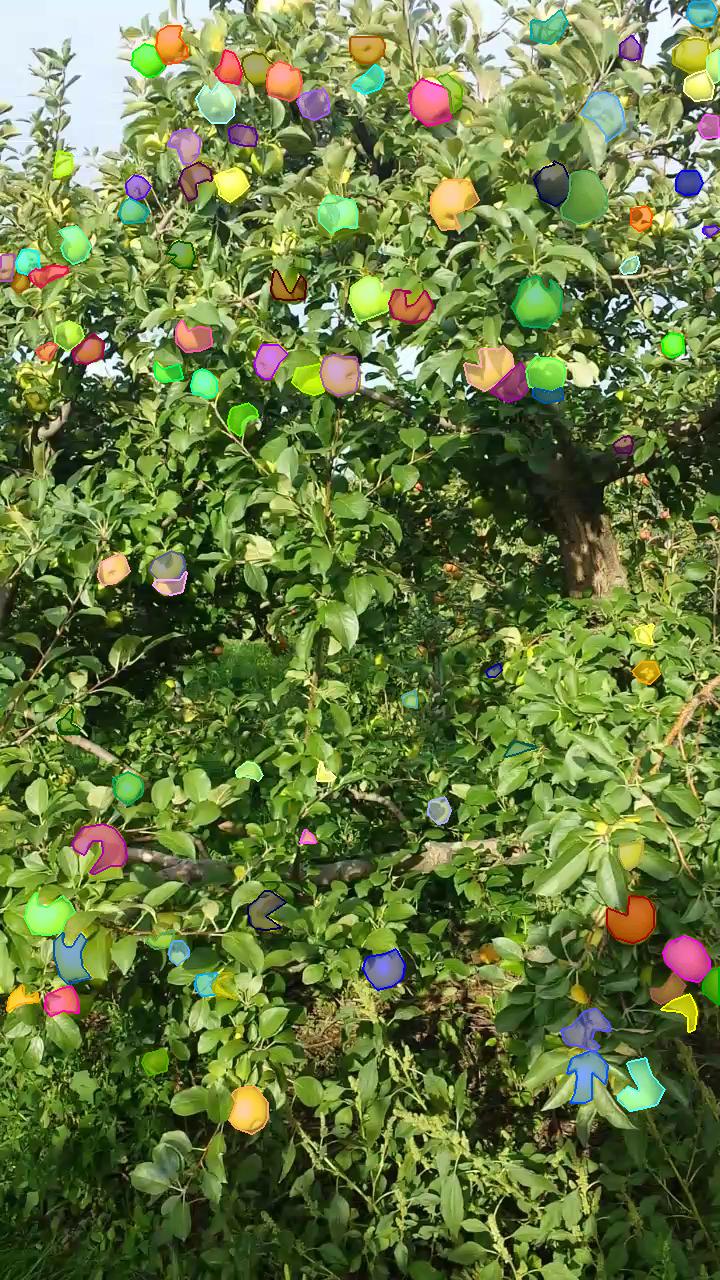}\\
{\scriptsize Input image} & {\scriptsize AttentionMask} & {\scriptsize AttentionMask$^{4}_{16}$} & {\scriptsize T-FastMask}& {\scriptsize T-AttentionMask} & {\scriptsize Ground truth}
\end{tabular}
\caption{Qualitative results of standard AttentionMask~\cite{WilmsFrintropACCV2018}, AttentionMask$^{4}_{16}$, Tiled FastMask (T-FastMask), and Tiled AttentionMask (T-AttentionMask) on two images from our MinneApple test dataset~\cite{hani2020minneapple}. Since FastMask without tiling cannot locate any apple, we omit the qualitative results here. Filled colored contours denote found objects, while not filled red contours denote missed objects. Note that only the best fitting proposal (highest IoU) is visualized per annotated object.
}
\label{fig:minneapple_results}
\end{figure}

The qualitative results in Fig.~\ref{fig:minneapple_results} show two examples from the MinneApple dataset. Comparing the results indicates that the proposed At\-ten\-tion\-Mask$_4^{16}$ and Tiled AttentionMask locate more apples than standard AttentionMask~(see green arrows). Other apples or leaves partially occlude many apples missed by standard AttentionMask. As a result, the apples are very small, making localization more difficult for standard AttentionMask. Examples of such settings are found around the central arrow in the upper image in Fig.~\ref{fig:minneapple_results} or around the lower arrow in the lower images in Fig.~\ref{fig:minneapple_results}. The proposed approaches mitigate the problem of the small apple size by  utilizing the additional pyramid level or the tiling framework. Similar to At\-ten\-tion\-Mask$_4^{16}$, Tiled FastMask locates more apples than standard AttentionMask. Still, Tiled FastMask misses several small or occluded apples that Tiled AttentionMask locates on the pyramid level $\mathcal{S}_8$. These qualitative results are in line with the quantitative results.

In general, it is well visible that several apples are difficult to localize, even for humans. This is due to the high complexity of the scenes with occlusions and many leaves. Additionally, some leaves have  similar colors and shapes compared to the apples, making localization even more complex. Overall, the qualitative results underline the strength of the proposed approaches in finding small apples and the high complexity of the data.

\section{Conclusion}
\label{sec:conclusion}
In this paper, we tackled the problem of localizing apples in complex apple orchard environments for agricultural applications. We specifically focused on small apples, as they usually dominate images depicting entire apple trees. For improved results on localizing such apples, we proposed two approaches based on the object proposal generation system AttentionMask. The first approach adds a new module for very small apples, while the second approach utilizes a tiling strategy. Both approaches clearly outperform standard AttentionMask and FastMask on the MinneApple dataset. The improvements are driven by the success in localizing very small apples due to the architectural changes. We also showed that both proposed approaches have specific advantages, focusing on efficient processing or overall localization quality. However, more research towards a better localization is necessary, as the quantitative results showed several missed apples.

\bibliographystyle{splncs04}
\bibliography{lit}

\begin{thebibliography}{10}
\providecommand{\url}[1]{\texttt{#1}}
\providecommand{\urlprefix}{URL }
\providecommand{\doi}[1]{https://doi.org/#1}

\bibitem{alexe2010object}
Alexe, B., Deselaers, T., Ferrari, V.: What is an object? In: Computer Vision
  and Pattern Recognition (2010)

\bibitem{anderson2019estimation}
Anderson, N.T., Underwood, J.P., Rahman, M.M., Robson, A.J., Walsh, K.B.:
  Estimation of fruit load in mango orchards: Tree sampling considerations and
  use of machine vision and satellite imagery. Precision Agriculture
  \textbf{20}(4) (2019)

\bibitem{bargoti2017deep}
Bargoti, S., Underwood, J.: Deep fruit detection in orchards. In: International
  Conference on Robotics and Automation (2017)

\bibitem{gongal2015sensors}
Gongal, A., Amatya, S., Karkee, M., Zhang, Q., Lewis, K.: Sensors and systems
  for fruit detection and localization: A review. Computers and Electronics in
  Agriculture  \textbf{116} (2015)

\bibitem{hani2020comparative}
H{\"a}ni, N., Roy, P., Isler, V.: A comparative study of fruit detection and
  counting methods for yield mapping in apple orchards. Journal of Field
  Robotics  \textbf{37}(2) (2020)

\bibitem{hani2020minneapple}
H{\"a}ni, N., Roy, P., Isler, V.: {MinneApple}: A benchmark dataset for apple
  detection and segmentation. IEEE Robotics and Automation Letters
  \textbf{5}(2) (2020)

\bibitem{he2017mask}
He, K., Gkioxari, G., Doll{\'a}r, P., Girshick, R.: Mask {R-CNN}. In:
  International Conference on Computer Vision (2017)

\bibitem{he2016deep}
He, K., Zhang, X., Ren, S., Sun, J.: Deep residual learning for image
  recognition. In: Computer Vision and Pattern Recognition (2016)

\bibitem{Hosang2015PAMI}
Hosang, J., Benenson, R., Doll\'{a}r, P., Schiele, B.: What makes for effective
  detection proposals? IEEE Transactions on Pattern Analysis and Machine
  Intelligence  \textbf{38}(4) (2015)

\bibitem{Hu2017-fastmask}
Hu, H., Lan, S., Jiang, Y., Cao, Z., Sha, F.: {FastMask}: Segment multi-scale
  object candidates in one shot. In: Computer Vision and Pattern Recognition
  (2017)

\bibitem{koirala2019deepReview}
Koirala, A., Walsh, K.B., Wang, Z., McCarthy, C.: Deep learning--method
  overview and review of use for fruit detection and yield estimation.
  Computers and Electronics in Agriculture  \textbf{162} (2019)

\bibitem{koirala2019deep}
Koirala, A., Walsh, K.B., Wang, Z., McCarthy, C.: Deep learning for real-time
  fruit detection and orchard fruit load estimation: Benchmarking of
  ‘{MangoYOLO}’. Precision Agriculture  \textbf{20}(6) (2019)

\bibitem{lin2014microsoft}
Lin, T.Y., Maire, M., Belongie, S., Hays, J., Perona, P., Ramanan, D.,
  Doll{\'a}r, P.: Microsoft {COCO}: Common objects in context. In: European
  Conference on Computer Vision (2014)

\bibitem{liu2020deep}
Liu, L., Ouyang, W., Wang, X., Fieguth, P., Chen, J., Liu, X., Pietik{\"a}inen,
  M.: Deep learning for generic object detection: A survey. International
  Journal of Computer Vision (IJCV)  \textbf{128}(2) (2020)

\bibitem{mai2018faster}
Mai, X., Zhang, H., Meng, M.Q.H.: {Faster R-CNN} with classifier fusion for
  small fruit detection. In: International Conference on Robotics and
  Automation (2018)

\bibitem{Pinheiro2015-deepmask}
Pinheiro, P.O., Collobert, R., Doll{\'a}r, P.: Learning to segment object
  candidates. In: Advances in Neural Information Processing Systems (2015)

\bibitem{Pinheiro2016-sharpmask}
Pinheiro, P.O., Lin, T.Y., Collobert, R., Doll{\'a}r, P.: Learning to refine
  object segments. In: European Conference on Computer Vision (2016)

\bibitem{redmon2016you}
Redmon, J., Divvala, S., Girshick, R., Farhadi, A.: You only look once:
  Unified, real-time object detection. In: Computer Vision and Pattern
  Recognition (2016)

\bibitem{ren2016faster}
Ren, S., He, K., Girshick, R., Sun, J.: {Faster R-CNN: T}owards real-time
  object detection with region proposal networks. IEEE Transactions on Pattern
  Analysis and Machine Intelligence  \textbf{39}(6) (2016)

\bibitem{sa2016deepfruits}
Sa, I., Ge, Z., Dayoub, F., Upcroft, B., Perez, T., McCool, C.: {DeepFruits}: A
  fruit detection system using deep neural networks. Sensors  \textbf{16}(8)
  (2016)

\bibitem{stein2016image}
Stein, M., Bargoti, S., Underwood, J.: Image based mango fruit detection,
  localisation and yield estimation using multiple view geometry. Sensors
  \textbf{16}(11) (2016)

\bibitem{wang2013automated}
Wang, Q., Nuske, S., Bergerman, M., Singh, S.: Automated crop yield estimation
  for apple orchards. In: International Symposium on Experimental Robotics
  (2013)

\bibitem{WilmsFrintropACCV2018}
Wilms, C., Frintrop, S.: {AttentionMask}: Attentive, efficient object proposal
  generation focusing on small objects. In: Asian Conference on Computer Vision
  (2018)

\bibitem{WilmsFrintropICPR2020}
Wilms, C., Frintrop, S.: Superpixel-based refinement for object proposal
  generation. In: International Conference on Pattern Recognition (2020)

\bibitem{WilmsFrintropIVC2021}
Wilms, C., Frintrop, S.: {DeepFH} segmentations for superpixel-based object
  proposal refinement. Image and Vision Computing  \textbf{114} (2021)

\bibitem{WilmsEtAlICPR2020}
Wilms, C., Heid, R., Sadeghi, M.A., Ribbrock, A., Frintrop, S.: Which airline
  is this? {A}irline logo detection in real-world weather conditions. In:
  International Conference on Pattern Recognition (2020)

\bibitem{yu2019fruit}
Yu, Y., Zhang, K., Yang, L., Zhang, D.: Fruit detection for strawberry
  harvesting robot in non-structural environment based on {Mask-RCNN}.
  Computers and Electronics in Agriculture  \textbf{163} (2019)

\end{thebibliography}
\end{document}